\documentclass[runningheads]{llncs}
\makeatletter
\newcommand{\printfnsymbol}[1]{%
  \textsuperscript{\@fnsymbol{#1}}%
}
\makeatother
\usepackage{graphicx}
\usepackage{array}
\usepackage{multirow}
\usepackage{graphicx}
\usepackage{comment}
\usepackage{amsmath,amssymb} 
\usepackage{color}
\begin{document}
\title{{Fast Point Voxel Convolution Neural Network with Selective Feature Fusion for Point Cloud Semantic Segmentation}}
\titlerunning{Fast Point Voxel Convolution}
%
%
%
 \author{Xu Wang\thanks{Equal contribution} \and
 Yuyan Li\printfnsymbol{1} \and
 Ye Duan}
 \authorrunning{Wang, Li. et al.}
%
 \institute{University of Missouri, USA \\
\email{\{xwf32, yl235, duanye\}@umsystem.edu}\\}

\maketitle              
\begin{abstract}
We present a novel lightweight convolutional neural network for point cloud analysis. In contrast to many current CNNs which increase receptive field by downsampling point cloud, our method directly operates on the entire point sets without sampling and achieves good performances efficiently. Our network consists of point voxel convolution (PVC) layer as building block. Each layer has two parallel branches, namely the voxel branch and the point branch. For the voxel branch specifically, we aggregate local features on non-empty voxel centers to reduce geometric information loss caused by voxelization, then apply volumetric convolutions to enhance local neighborhood geometry encoding. For the point branch, we use Multi-Layer Perceptron (MLP) to extract fine-detailed point-wise features. Outputs from these two branches are adaptively fused via a feature selection module. Moreover, we supervise the output from every PVC layer to learn different levels of semantic information. The final prediction is made by averaging all intermediate predictions. We demonstrate empirically that our method is able to achieve comparable results while being fast and memory efficient. We evaluate our method on popular point cloud datasets for object classification and semantic segmentation tasks.

\end{abstract}
\section{Introduction}

Deep learning in 3D point cloud analysis has received increasing attention with the rising trend of Virtual Reality and 3D scene understanding applications, etc. Existing approaches have made great progresses in tasks such as point cloud classification \cite{wu20153d} and point cloud semantic segmentation \cite{chang2015shapenet,armeni2017joint}. One fundamental issue to be tackled with in point cloud analysis is the representation of unstructured point clouds. Some early methods discretize point clouds into regular volumetric grids which can be directly fed into standard 3D CNNs. However, two main problems coupled with this volumetric representation are information loss and huge memory consumption. A high resolution voxel grid leads to expensive computation cost, while a low resolution inevitably suffers from information loss during voxelization procedure. 

\begin{figure}[t]
    \centering
    \includegraphics[width=9cm]{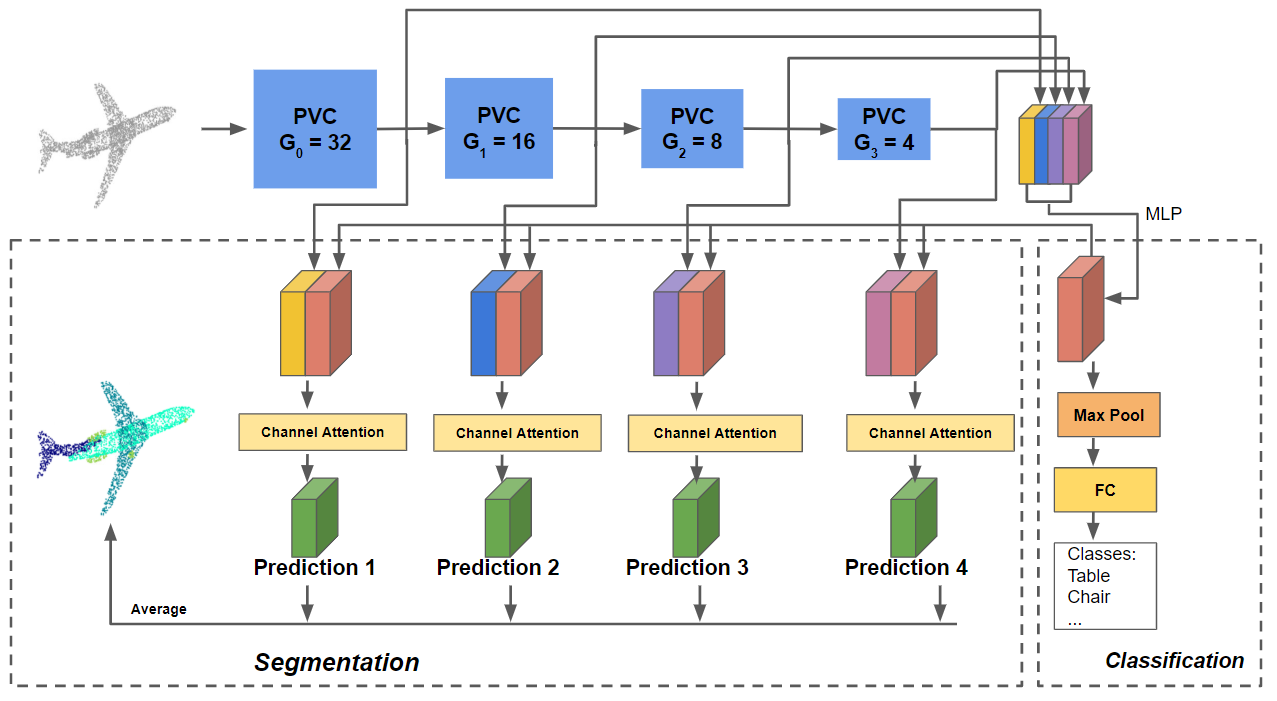}
    \caption{Illustration of our proposed network. For an input 3D data, we pass it through a sequence of point-voxel convolution layers(PVC). $G$ denotes grid resolution. Outputs from each PVC layer are concatenated together to form a global feature. MLP is used for feature dimension reduction. This global feature is concatenated with output from each PVC layer and passed through a channel attention module, which re-weights the features of all channels and increases feature disciminability. The final prediction is the average of all auxiliary predictions.}
    \label{fig:pvnetwork}
\end{figure}

To address the problems mentioned above, another big stream is to directly consume sparse point clouds. The pioneer work is PointNet proposed by Qi et al. \cite{qi2017pointnet}.  PointNet is able to process unordered point cloud inputs with permutation invariance using a sequence of multi-layer perceptron(MLP). The subsequent work PointNet++\cite{qi2017pointnet++} achieves better performance by proposing a hierarchical network that encodes local neighborhood information. Based on PointNet++\cite{qi2017pointnet++}, a great number of networks\cite{zhao2019pointweb,zhang2019shellnet,wang2019graph} with more advanced local feature aggregation techniques are introduced. Apart from MLP-based methods, some recent works propose kernel-based approaches to mimic standard convolution\cite{thomas2019kpconv,wang2018deep,xu2018spidercnn,boulch2020convpoint}. In general, point-based approaches suffer from point sampling scalability, neighbor point searching efficiency, point density inconsistency issues. Most recently, Liu et al. \cite{liu2019point} attempt to design a point-voxel CNN that represents 3D data as points to reduce memory footprint, and leverages voxel-based convolution to capture neighborhood features. This network is able to achieve reasonable performance with low memory usage and fast training/inference speed.  

In this paper, we propose a novel CNN architecture that is well-balanced between efficiency and accuracy. Inspired by point-voxel CNN\cite{liu2019point}, we construct our network using point-voxel layer that takes advantages of both sparse point representation and volumetric convolution. Our point-voxel layer consists of two parallel branches, a voxel-based branch which aggregates local neighboring features, and a point-based branch which maintains fine-grained point-wise features.  During discretization in voxel branch, we aggregate neighboring features on non-empty voxel centers and use standard 3D convolutions to enhance local feature encoding. Voxel features are propagated back to point domain through devoxelization.  Outputs from point and voxel branches are fused self-adaptively via a feature selection module(FSM), which learns channel-wise attention for both branches.

Most of the existing studies rely heavily on point sampling strategy to avoid expensive computation cost as network goes deeper. However, point sampling cannot always retain the fine-detailed features for every point. Details of points are discarded as a trade-off for larger receptive field and processing speed.
In our network, we only use a small number of point-voxel layers (default is 4 layers) that are carefully designed with effective feature encoding modules to facilitate processing efficiency. Supervision is applied on outputs from all layers to enforce semantic information learning. 
Though no point sampling is conducted, our network is able to remain lightweight, and effective to process large-scale point clouds.
A visualization of our proposed network is shown in Figure \ref{fig:pvnetwork}.
We evaluate the performance and efficiency of our proposed model for object classification, object part segmentation and indoor scene semantic segmentation tasks (see Section 4).

\section{Related Work}
\subsection{Volumentric Representation}
Some early deep learning approaches transformed point clouds into 3D voxel structure and convolve it with standard 3D kernels.  VoxNet \cite{maturana2015voxnet} and subsequent works \cite{wu20153d,maturana20153d,tchapmi2017segcloud} discretized point cloud into a 3D binary occupancy grid. The occupancy grid is fed to a CNN for object proposal and classification. These voxel-based methods suffered from high memory consumption due to the waste of computation on empty spaces. OctNet\cite{riegler2017octnet,tatarchenko2017octree} proposed adaptive representation using octree structure to reduce memory consumption. Recent researches \cite{graham20183d,choy20194d} introduced approaches to process high dimensional data and apply sparse convolution only on non-empty voxels. In general, volumetric methods preserve neighborhood information of point clouds, enable regular 3D CNN applications, but suffer from significant discretization artifacts. 

\subsection{Point-based Representation}
Point-wise models such as PointNet \cite{qi2017pointnet} and PointNet++ \cite{qi2017pointnet++} directly operates on point clouds. The former used MLPs to extract point-wise features and permutation- invariant max pooling operation to obtain a global feature. The latter built a hierarchical architecture that incorporates point downsampling and local structure aggregation strategies. Inspired by PointNet \cite{qi2017pointnet} and PointNet++ \cite{qi2017pointnet++}, many recent works propose advanced local feature learning modules. For example, PointWeb \cite{zhao2019pointweb} built a dense fully connected web to explore local context, and used an Adaptive Feature Adjustment module for feature refinement. GACNet \cite{wang2019graph} proposed to selectively learn distinctive features by dynamically assigning attention weights to neighbouring points based on spatial positions and feature differences.
ShellNet \cite{zhang2019shellnet} built a model with several layers of ShellConv, and solved point ambiguity by constructing concentric shells and applying 1D convolution on ordered shells.
Derived from point-based methods, some recent works define explicit kernels for point convolution. KCNet \cite{shen2018mining} developed a kernel correlation layer to compute affinities between each point’s K nearest neighbors and a predefined set of kernel points. Local features are acquired by graph pooling layers. SpiderCNN \cite{xu2018spidercnn} designed a family of Taylor polynomial kernels to aggregate neighbor features. PointCNN \cite{li2018pointcnn} introduced $\chi$-transformation to exploit the canonical order of points. PCNN \cite{wang2018deep} built a network using parametric continuous convolutional layers. SPH3D \cite{lei2019spherical} used spherical harmonic kernels during convolution on quantized space to identify distinctive geometric features. KPConv \cite{thomas2019kpconv} defined rigid and deformable kernel points for local geometry encoding based on the Euclidean space relations between kernel point and neighborhood supporting points.

\subsection{Efficiency of Current Models}
When processing large-scale point clouds, efficiency is one of the fundamental measurements to evaluate models. Most of the point-based methods utilized point sampling to improve efficiency. However, it is non-trivial to choose an effective point sampling method.  For example, Farthest Point Sampling(FPS) which is widely adopted in \cite{qi2017pointnet++,zhao2019pointweb,zhang2019shellnet}, has $O(NlogN)$ computation complexity, meaning it does not have good scalability. Random point sampling as used in RandLA-Net\cite{hu2019randla}, has $O(1)$ time complexity, but random point sampling cannot be invariant to point densities and key information might be discarded. Other approaches manage to incorporate hybrid representations to avoid the redundant computing and storing of more useful spatial information. A recent work Grid-GCN \cite{xu2019grid} proposed a novel method which facilitates grid space structuring and provides more complete coverage of the point cloud. This method is able to handle massive points with fast speed and good scalability.
Point-Voxel CNN \cite{liu2019point} is most related to our method. This work combines fine-grained point features with coarse-grained voxel features with speedup and low memory consumption. Compared to their work, our method builds PVC layers with multi-resolution voxels, and incorporates more accurate local feature aggregation which reduce information loss artifacts. 

\section{Method}
We build a deep architecture with a sequence of point-voxel convolution(PVC) layers. In this section, we introduce the details of our PVC layer, including voxelization, local feature aggregation, devoxelization, selective feature fusion, and deep supervision.

\subsection{Point Voxel Convolution}
\subsubsection{Voxelization and Local Aggregation}
The purpose of voxel branch is to encode contextual information through volumetric convolution. As aforementioned, information loss from the process of discretization is inevitable. Introducing large voxel grids reduces the loss, but burdens the network with huge computation overhead. In our design, we opt to use low-resolution volumetric grid, and mitigate information loss by effective local feature aggregation. 

As the scale of point clouds varies, we first normalize the input data into a bounding box. 
For voxelization, we quantize point cloud by calculating voxel coordinates $(u,v,w)\in \mathbb{N}$ from point coordinates$(x,y,z)\in \mathbb{R}$.
\begin{equation}
    u = floor((x-x_{min})/g_x),\; v = floor((y-y_{min})/g_y),\; w = floor((z-z_{min})/g_z)
\end{equation}
where $g_x, g_y, g_z$ is the grid length of $x,y,z$ axis respectively: 
\begin{equation}
    g_x = (x_{max}-x_{min})/G,\; g_y = (y_{max}-y_{min})/G,\; g_z = (z_{max}-z_{min})/G
\end{equation}
where $G\in \mathbb{N}$ is the grid resolution of this PVC layer.


Given the voxelized point cloud, we calculate the center location $P_{c}$ of every voxel. To accelerate processing speed, we locate non-empty voxels and aggregates local features only on these voxels. For a non-empty voxel $(u,v,w)$, we use the cell center as query position and gather K neighbors through K-nearest-neighbor(KNN).
We adopt the dilated point convolution strategy as proposed in \cite{engelmann2019dilated}, in which $K\cdot n$ nearest neighbors and every $n-th$ neighbor is selected. The feature $f$ for voxel $(u,v,w)$ is the weighted summation of all K neighbors.  
\begin{equation}
    f_{u,v,w} = \sum_k^K l_k * f_k
\end{equation}
where $l_k\in \mathbb{R}^{1\times C}$ is the weight, $f_k\in \mathbb{R}^{1\times C}$ is the feature of k-th neighbor. Inspired by \cite{wang2019graph,wang2018deep}, we use self-attention mechanism to learn weight of different neighboring points:
\begin{equation}
    l_k = \sigma(\mathcal{G}(\Delta_{P_{k}}, f_k)) 
\end{equation}
where $\Delta_{P_{k}}$ is the normalized neighbor point coordinates, $\Delta_{P_{k}} = P_{k}-P_c$. $\mathcal{G}(*)$ takes the concatenation of  $\Delta_{P_{k}}$ and $f_k$ as input and models the attention weight. $\sigma$ represents ReLU activation function. Through neighbor feature aggregation, we collect useful information and store it in the voxels. Next, we apply 3D convolutions on the voxel grid, as an enhancement of local neighborhood learning. 

\subsubsection{Devoxelization}
To allow feature fusion from two branches, we propagate voxel features back to point domain based on their voxel coordinates. Taking efficiency into account, we assign voxel feature $f_(u,v,w)$ to all points that fall into this voxel. We observe that in our experiment since point-wise features are carried all along by point branch, fusing voxel feature with point feature would be effective to discriminate individual point.

\subsubsection{Point and Voxel Feature Fusion}
\begin{figure}[t!]
    \centering
    \includegraphics[width=10cm]{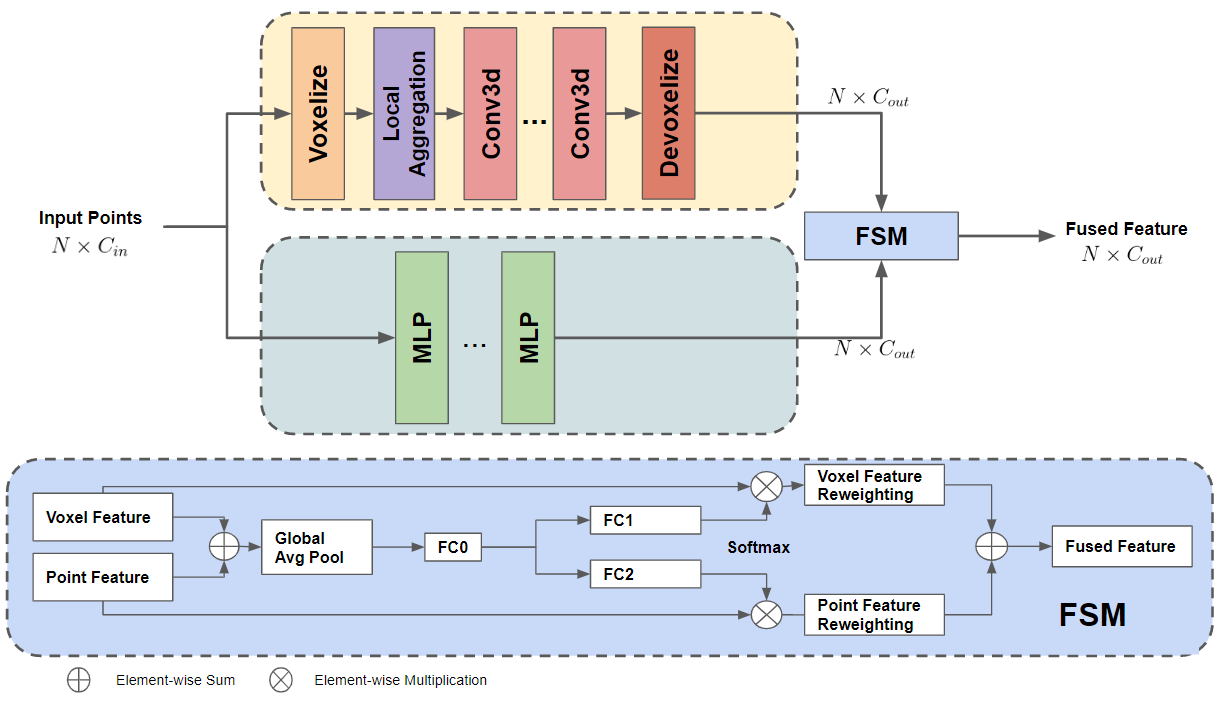}
    \caption{Illustration of our PVC layer (top) and Feature Selection Module (bottom). }
    \label{fig:pvlayer}
\end{figure}

As illustrated in Figure \ref{fig:pvlayer}, for point branch, we use MLPs to extract point-wise features. While voxel features encode local neighborhood, point branch is able to carry fine-detailed per-point features.
Next, we incorporate a feature selection module to correlate features. First, we use element-wise summation to fuse features from point and voxel branches:
\begin{equation}
    f' = f_p + f_v
\end{equation}
where $f_p\in \mathbb{R}^{N\times C}$ and $f_v\in \mathbb{R}^{N\times C}$ are features from point and voxel branch respectively. Then a global average pooling is applied to squeeze $N$ point to one compact point feature. Fully connected layer with non-linearity is used to provide guidance for feature selection:
\begin{equation}
    S = \sigma(\mathcal{F}_{gp}(f')\cdot W_{fc})
\end{equation}
where $\sigma$ is the ReLU activation function, $\mathcal{F}_{gp}$ is the global average pooling, $W_{fc}\in \mathbb{R}^{C\times d}(d=C/4)$ is the learnable weight for fully connected layer. 
Two separate fully connected layers are applied to increase channel dimensions for $S$ and produce soft attention vector $S_p\in \mathbb{R}^{1\times C}$ and $S_v\in \mathbb{R}^{1\times C}$. 
\begin{equation}
    S_p = S\cdot W_1, \; \; S_v = S\cdot W_2
\end{equation}
where $W_1\in \mathbb{R}^{d\times C}$ and $W_2\in \mathbb{R}^{d\times C}$ are learnable weights. We adopt the softmax mechanism on $S_p$ and $S_v$ to adaptively select features.
\begin{equation}
    S_{p,c} = \frac{e^{S_p,c}}{e^{S_p,c}+e^{S_v,c}},\; S_{v,c} = \frac{e^{S_v,c}}{e^{S_p,c}+e^{S_v,c}}
\end{equation}
where $S_{p,c}$ and $S_{v,c}$ are soft attention vector for point and voxel feature at $c^{th}$ channel.
The fused feature at $c^{th}$ channel can be calculated as follows:
\begin{equation}
    F_{fused, c} = S_{p,c} \odot F_{p,c} + S_{v,c} \odot F_{v,c}
\end{equation}
Therefore, FSM adjusts channel-wise weight for different branches, and outputs the fused feature adaptively.

\subsection{Deep Supervision}
As shown in Figure \ref{fig:pvnetwork}, we build our network with several PVC layers sequentially. Grid resolution decreases while output feature channel increases from shallow to deep. Different PVC layers extract different levels of semantic information. We add supervision on each PVC layer output to enforce different levels of semantic feature learning. Similar strategy is adopted by \cite{sinha2019multi} for multi-scale medical image segmentation. In detail, we concatenate outputs from all PVC layers and use MLPs to produce a compact feature. Serving as a global guidance, this feature map is concatenated with the output of each PVC layer, then pass through a channel attention module to enhance feature representation of specific semantics. This attention module, inspired by \cite{fu2019dual}, aggregates weighted features of all the channels into the original features, and models discriminability between channels. The output from channel attention module produce an auxiliary loss. We add up all the losses and average the prediction probabilities for the final prediction. We show that incorporating channel attention module in our network boosts performances and not necessarily slow down inference speed(see Section 4.4).
 \begin{figure}
     \centering
     \includegraphics[width=7cm]{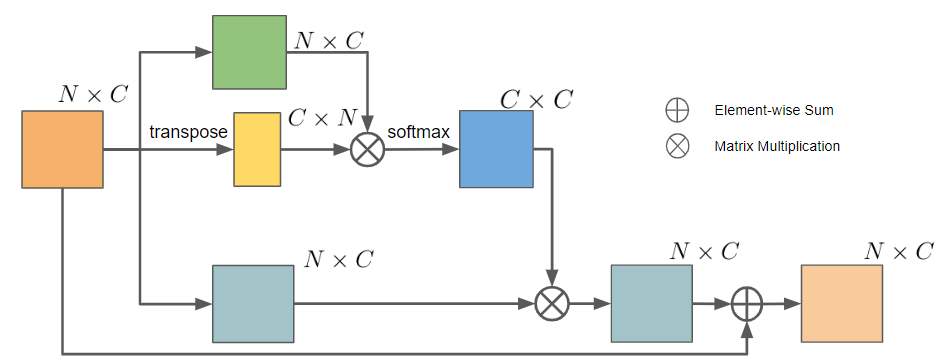}
     \caption{Channel attention module}
     \label{fig:channel_att}
     \vspace{-0.75cm}
 \end{figure}

\section{Experiment}
In this section, we evaluate the performance of our method in point cloud learning tasks including object classification, object part segmentation, and indoor scene segmentation. For parameter settings, our network has a total of four PVC layers. From the first to the last layer, grid size is set to $G_0 = 32, G_1=16, G_2=8, G_3=4$ respectively. Number of neighbors for KNN search is 32. Dilation step is $n_0=1, n_1=2, n_2=4, n_3=8$. Our method is implemented with PyTorch \cite{paszke2017automatic} and run on a Nvidia TitanXP GPU.
Batch size for training is set to be 16. We use the Adam optimizer \cite{kingma2014adam} with default settings. The learning rate is initialized as 0.001 and decays by a rate of 0.5 every 20 epochs. Object classification converges around 100 epochs, part segmentation converges at 80 epochs, and indoor scene segmentation converges at 120 epochs. The full version of our network has $C=64$ feature channels for the first PVC layer. Feature channel for l-th layer is $C\times 2^{l-1}$.

\subsection{Object Classification}
\subsubsection{Datasets and Implementation Details}
We evaluate our network on ModelNet40 \cite{wu20153d} for 3D object classification task. ModelNet40 contains 12311 meshed CAD models from 40 categories. The dataset has 9843 objects for training and 2468 objects for testing. We prepare the dataset following PointNet \cite{qi2017pointnet} conventions. We random sample $N$ points and only use normal as input feature. As illustrated in Figure \ref{fig:pvnetwork}, the output from all PVC layers are concatenated to form a global feature. Max pooling followed by a fully connected layer are then used to produce a classification score.

\subsubsection{Evaluation}
We train three versions of classification network with variance on number of points and number of feature channels. Shown in Table \ref{tab:modelnet40}, our method has a good balance between performance and efficiency.

\begin{table}[tbh!]
\caption{Results of ModelNet40 \cite{wu20153d} classification. $1\times C$ represents our full size model.}
  \centering
  \begin{footnotesize}
    \begin{tabular}{lccc}
    \hline
        Method &  Input Data  & OA  & Latency(ms) \\
        \hline \hline
        PointNet \cite{qi2017pointnet} & $8\times1024$ & 89.2 & $\mathbf{15}$ \\
         Ours($0.5\times C$) & $8\times1024$ & $\mathbf{91.7}$ & 20\\
         \hline
        PointNet++ \cite{qi2017pointnet++} & $8\times1024$ &91.9 & 27 \\
        DGCNN \cite{wang2019dynamic} & $8\times1024$ & 91.9 & 27 \\
        Ours($0.75\times C$) & $8\times1024$ & $\mathbf{92.3}$ & $\mathbf{26}$\\
        \hline
        Grid-GCN(full) \cite{xu2019grid} & $16\times1024$ & $ \mathbf{93.1}$ & 42 \\
        Ours($1\times C$) & $8\times2048$ & 92.5 & $\mathbf{35}$\\
        \hline 
    \end{tabular}
    \label{tab:modelnet40}
    \end{footnotesize}
    \vspace{-.5cm}
\end{table}

\subsection{Shape Segmentation}
\subsubsection{Data and Implementation Details}We conduct experiment on ShapeNetPart\cite{chang2015shapenet} for shape segmentation. ShapeNetPart is a collection of 16681 point clouds (14006 for training, 2874 for testing) from 16 categories, each annotated with 2-6 labels. Input features are normals only, while point coordinates $(x,y,z)$ are incorporated in the network for voxelization and local aggregation. We random sample 2048 points for training and use the original points for testing.  
\subsubsection{Evaluation}
A comparison of our method and previous approaches is listed in Table \ref{tab:shapenet_speed}. We report our result with mean instance IoU. We train three versions of our method, a compact network with $0.5\times C$ feature channels, a medium size network with $0.75\times C$ feature channels, and a full size network. Our compact network is able to achieve the same results as PointNet++\cite{qi2017pointnet++}, with $2\times$ speedup, and 0.7G less memory consumption. Comparing with DGCNN our compact method is $2\times$ faster and only half of its memory usage. Comparing with PV CNN\cite{liu2019point}, our method achieves comparable results.

\begin{table}[tbh!]
\caption{Results of object part segmentation on ShapeNetPart \cite{chang2015shapenet}. Our method achieves comparable performance with fast inference speed, and low GPU consumption.}
\centering
\begin{footnotesize}
    \begin{tabular}{l|c|c|c|c}
    \hline
        Method &  Input Data & InstanceIoU  & Latency (ms) & GPU usage (G)\\
        \hline
        \hline
         PointNet \cite{qi2017pointnet} & $8\times 2048$ & 83.7 & $\mathbf{22}$ & 1.5\\
         3D-Unet \cite{cciccek20163d} & Volume($8\times 96^3$) & 84.6 & 682 & 8.8\\
         PointNet++ \cite{qi2017pointnet++} & $8\times 2048$ & 85.1 & 78 & 2.0\\
         DGCNN \cite{wang2019dynamic} & $8\times 2048$ & 85.1 & 88 & 2.4\\
         PV CNN($0.5\times C$) \cite{liu2019point} & $8\times 2048$ & $\mathbf{85.5}$ & 22 & $\mathbf{1.0}$\\
          Ours($0.5\times C$) & $8\times 2048$ & $\mathbf{85.5}$ & 32 & 1.3\\
          \hline
         PointCNN \cite{li2018pointcnn} & $8\times 2048$ & 86.1 & 136 & 2.5\\
         PV CNN($1\times C$) \cite{liu2019point} & $8\times 2048$ & 86.2 & $\mathbf{51}$ & $\mathbf{1.6}$\\
         Ours($1\times C$) & $8\times 2048$ & $\mathbf{86.3}$ & 68 & 2.3\\
         \hline
    \end{tabular}
    \label{tab:shapenet_speed}
    \end{footnotesize}
    \vspace{-0.5cm}
\end{table}

\subsection{Indoor Scene Segmentation}
\subsubsection{Data and Implementation Details}
We conduct experiments on S3DIS \cite{armeni2017joint} for large-scale indoor scene segmentation. S3DIS \cite{armeni2017joint} is a challenging dataset which consists of point clouds collected from six areas. Following the convention \cite{qi2017pointnet,li2018pointcnn}, we leave out area 5 for testing purpose. For data preparation, we split rooms into $2m\times 2m$ blocks, with $0.5m$ padding along each side $(x,y)$. These context points do not involve in neither loss computation nor prediction during testing. We use color as input feature, point coordinates are incorporated in the network for voxelization and local aggregation. At training time, we random sample 8192 points from block data, and use original points at testing time. 

To demonstrate the great potential of our proposed network, we also design experiments which replace regular 3D convolution layers with sparse 3D convolutions. We adopt Minkowski \cite{choy20194d} sparse convolution in our experiment.  Sparse convolution enables our network to process high-dimensional data with further speedup and reduce computation load on GPU.
\subsubsection{Evaluation}
A list of comparison of our method and previous approaches is shown in Table \ref{tab:s3d_speed}. We also train three versions for indoor scene segmentation. Compared with PV-CNN++ \cite{liu2019point}, our method is faster while able to achieve $2.9\%$ higher mIoU score. 

\begin{table}[t]
\caption{Results of indoor scene segmentation on S3DIS\cite{armeni2017joint}, evaluated on Area 5. We report the result using mean Intersection-Over-Union(mIoU) metric. Compared with previous methods, our method is able to achieve top-ranking results while being lightweight and fast.}
\begin{footnotesize}
\centering
    \begin{tabular}{l|c|c|c|c}
    \hline
        Method &  Input Data & mIoU  & Latency (ms) & GPU usage (G)\\
        \hline
        \hline
         PointNet \cite{qi2017pointnet} & $8\times 4096$ & 43.0 & $\mathbf{21}$ & 1.0\\
         PointNet++ \cite{qi2017pointnet++} & $8\times 4096$ & 52.3 & - & -\\
         3D-Unet \cite{cciccek20163d} & Volume($8\times 96^3$) & 55.0 & 575 & 6.8\\
         DGCNN \cite{wang2019dynamic} & $8\times 4096$ & 48.0 & 178 & 2.4\\
         PointCNN \cite{li2018pointcnn} & $16\times 2048$ & 57.3 & 282 & 4.6\\
         PV CNN++($0.5\times C$) \cite{liu2019point} & $4\times 8192$ &57.6 & 41 & $\mathbf{0.7}$\\
         PV CNN++($1\times C$) \cite{liu2019point} & $4\times 8192$ & 59.0 & 70 & 0.8\\
         Grid-GCN(full) \cite{xu2019grid} & $8\times 4096$ & 57.8 & 26 & - \\
         \hline
         Ours($0.5\times C$) & $4\times 8192$ & 60.2 & 34 & 1.4\\
         Ours($0.75\times C$) & $4\times 8192$ & 60.8 & 51 & 1.8\\
         Ours($1\times C$) & $4\times 8192$ & $\mathbf{61.7}$ & 71 & 2.0\\
          Ours($1\times C$, sparse) & $4\times 8192$ & 61.4 & 42 & 0.9\\
         \hline
    \end{tabular}
    \label{tab:s3d_speed}
    \end{footnotesize}
\end{table}

\subsection{Ablation Study}
To show the effectiveness of our proposed method, we gradually add a component while keeping the rest unchanged. To see the gain of each component, we train a baseline network consists of four PVC layers. The baseline does not use local feature aggregation, instead an averaged feature of all points fall into the same voxel is taken. Fusion method is summation only. And no channel attention module(CAM) is used for prediction. Final prediction is directly produced from global feature without deep supervision. Experiments are conducted on ShapeNetPart\cite{chang2015shapenet}. The baseline model is the compact version ($0.5\times C$). From Table \ref{tab:ablation}, we can see that each component is able to boost baseline method without necessarily increase latency too much. 
\begin{table}[hbt!]
\caption{Ablation studies on ShapeNetPart\cite{chang2015shapenet}.}
\centering
\begin{footnotesize}
    \begin{tabular}{l|ccc}
        \hline
          &  mIoU & gain & Latency(ms)\\
        \hline
        Baseline &  84.6  &  - &  26\\
        w/ Local Aggregation & 84.8  &  +0.2 &  26\\
        w/ FSM & 85.2   &  +0.6 & 28  \\
         w/ CAM   & 84.9  &  +0.3 & 30\\
         Full(Local Aggregation + FSM + CAM) & 85.5  &  +0.9 & 32\\
         \hline
    \end{tabular}
    \label{tab:ablation}
    \end{footnotesize}
    \vspace{-.25cm}
\end{table}


\section{Conclusion}
In this work, we propose a novel approach for fast and effective 3D point cloud learning. We designed a lightweight network that can incorporate both fine-grained point features and multi-scale local neighborhood information. We introduce feature selection module and deep supervision into our network for performance improvement. Experimental results on several point cloud datsets demonstrate that our method achieves comparable results while being fast and memory efficient. 

\bibliographystyle{splncs04}
\bibliography{egbib}
\end{document}